\documentclass[letterpaper, 10 pt, conference]{ieeeconf}
\IEEEoverridecommandlockouts

\usepackage{cite}
\usepackage{graphics} 
\usepackage{graphicx}
\usepackage{subfig}
\usepackage{hyperref}
\hypersetup{
    colorlinks=true,
    linkcolor=blue,
    filecolor=magenta,      
    urlcolor=cyan,
    pdftitle={Overleaf Example},
    pdfpagemode=FullScreen,
    }

\usepackage{xcolor}

\usepackage{algorithm,algpseudocode}

\usepackage{maegan_style}

\title{POLAR$^1$: Preference Optimization and Learning Algorithms for Robotics}

\author{Maegan Tucker, Kejun Li, Yisong Yue, and Aaron D. Ames
\thanks{M. Tucker and A. Ames, are with the Department
of Mechanical and Civil Engineering, California Institute of Technology, Pasadena,
CA, 91125 USA. e-mail: \texttt{mtucker@caltech.edu.}}
\thanks{Y. Yue and A. Ames are with the Department of Computing and Mathematical Sciences, California Institute of Technology.}
\thanks{Y. Yue is with Argo AI.}
\thanks{K. Li is with the Department of Computation and Neural Systems, California Institute of Technology.}}

\begin{document}

\maketitle
\thispagestyle{empty}
\pagestyle{plain}
\begin{figure*}[t]
    \centering
    \includegraphics[width=\linewidth]{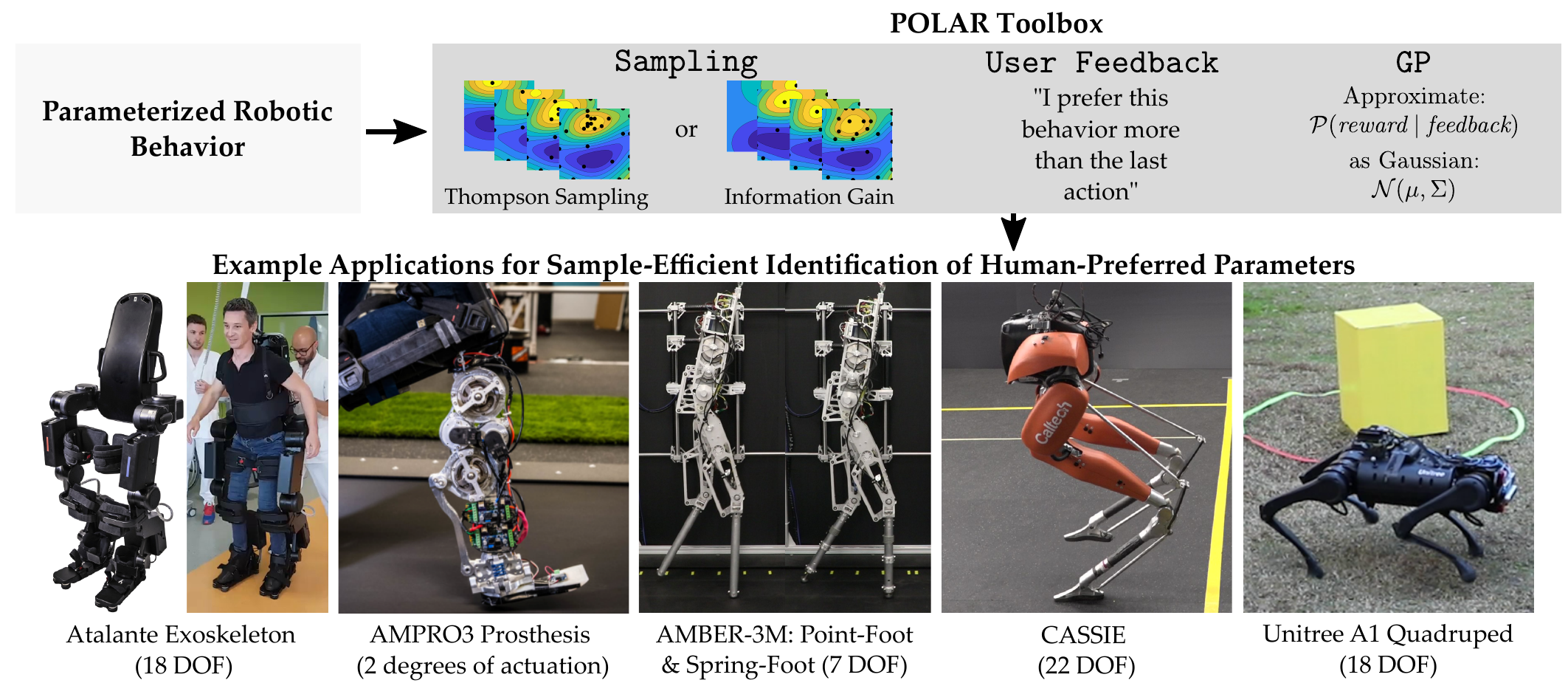}
    \caption{Given a parameterized robotic behavior, the POLAR toolbox executes human-in-the-loop preference-based learning framework that sequentially 1) samples new actions using either regret minimization or information gain, 2) queries the human for three forms of subjective feedback (pairwise preferences, user suggestions, and ordinal labels), and 3) approximates the underlying reward function using Gaussian processes. Preliminary versions of the POLAR toolbox was leveraged for the shown robotic platforms to identify human-preferred walking gaits and human-preferred control parameters.}
    \label{fig: intro-platforms}
\end{figure*}

\begin{abstract} 
Parameter tuning for robotic systems is a time-consuming and challenging task that often relies on domain expertise of the human operator. Moreover, existing learning methods are not well suited for parameter tuning for many reasons including: the absence of a clear numerical metric for `good robotic behavior'; limited data due to the reliance on real-world experimental data; and the large search space of parameter combinations. In this work, we present an open-source MATLAB Preference Optimization and Learning Algorithms for Robotics toolbox (POLAR) for systematically exploring high-dimensional parameter spaces using human-in-the-loop preference-based learning. This aim of this toolbox is to systematically and efficiently accomplish one of two objectives: 1) to \textit{optimize} robotic behaviors for human operator preference; 2) to learn the operator's underlying preference landscape to better \textit{understand} the relationship between adjustable parameters and operator preference. The POLAR toolbox achieves these objectives using only subjective feedback mechanisms (pairwise preferences, coactive feedback, and ordinal labels) to infer a Bayesian posterior over the underlying reward function dictating the user's preferences. We demonstrate the performance of the toolbox in simulation and present various applications of human-in-the-loop preference-based learning.

\end{abstract}


%
\IEEEpeerreviewmaketitle

\section{Introduction}
\label{Sec: Intro}


Achieving complex tasks on robotic systems often relies on heuristic tuning aimed at optimizing for holistic metrics such as `smoothness' or `naturalness'. Typically, this tuning process is labor-intensive and requires expert domain knowledge since the mapping between the adjustable parameters and the resulting robotic behavior is non-intuitive. To alleviate the tuning process and reduce the dependence on domain expertise, we propose the use of \textit{human-in-the-loop preference-based learning}, which sequentially selects parameter combinations to execute on the hardware and queries the human operator for subjective feedback. 

Recent work has demonstrated several successful applications of human-in-the-loop preference-based learning towards quickly identifying human-preferred robotic behaviors, without needing to either explicitly define the underlying reward function or numerically quantify \textit{good} behavior. These applications include achieving stable and robust bipedal locomotion \cite{csomay2021learning,tucker2021hzd} and identifying subject-preferred assistive device parameters \cite{tucker2020preference, ingraham2022role}.

This work introduces an open-source MATLAB toolbox \textit{Preference Optimization and Learning Algorithms for Robotics} (POLAR)
\footnote{POLAR toolbox: \url{https://github.com/maegant/POLAR} \\ Documentation: \url{https://maegant.github.io/POLAR/}} 
for easy deployment of  human-in-the-loop preference-based learning framework. As illustrated in Fig. \ref{fig: intro-platforms}, the POLAR toolbox can be applied to any parameterized robotic behavior and obtains either a human-preferred robotic behavior or the entire underlying preference landscape, depending on the selected sampling strategy. The main advantages of POLAR include that it 1) provides a unified treatment of previously presented preference-based learning algorithms, and 2) has various user-defined settings that allow the framework to be easily tailored to a wide variety of robotic applications.

Throughout the following sections of this paper we will present the details of the POLAR toolbox and demonstrate its capabilities. First, in Section \ref{sec: overview}, we present an overview of the problem statement and the unified POLAR architecture. The following three sections then detail the three overarching algorithm components: querying the human for three forms of subjective feedback (Section \ref{Sec: feedback}); approximating the underling reward function as a Gaussian process (Section \ref{Sec: modeling}); and using either a regret minimization or information gain sampling technique to select new actions to execute in the subsequent iteration (Section \ref{Sec: sampling}). After discussing these details, we present and discuss various simulation results (Section \ref{Sec: simulation}), followed by some case studies demonstrating potential uses of the toolbox (Section \ref{Sec: examples}). 

\subsection{Related Work}

 Three learning algorithms were introduced in prior work to individually accomplish the following: CoSpar for regret minimization \cite{tucker2020preference}; LineCoSpar for regret minimization in high-dimensional action spaces \cite{tucker2020human}; and ROIAL for information gain constrained to a \textit{region of interest} \cite{li2020roial}. The POLAR toolbox provides a unified treatment of these previously disjoint learning algorithms and extends the algorithms to leverage all three forms of subjective user feedback. 


Additionally, B\i y\i k et. al, recently released a python-based toolbox, APReL \cite{biyik2021aprel}, which features various sampling techniques aimed at active preference-based reward learning. 
In comparison to POLAR, which models the learning problem using Gaussian processes with new samples selected via either regret minimization or information gain, APReL formulates the learning problem as a discrete-time Markov Decision Process (MPD). In this setting, the learning objective is to identify a trajectory, defined as a sequence of state-action pairs, that maximizes the expected cumulative reward. Whereas the learning objective implemented by POLAR is to identify constant tuples of parameters that either minimize regret or characterize the relationship between the reward function and these parameters. 

\section{Preference-Based Learning Overview}
\label{sec: overview}

\subsection{Problem Setting}
Consider a robotic system with $\d \in \R$ adjustable parameters. The possible range of each parameter is assumed to be upper and lower bounded, with these bounds defined as $a^\text{max}_i, a^\text{min}_i \in \R$ respectively for each parameter $i = \{1,\dots,\d\}$. Also, since humans cannot easily distinguish between parameters with very similar values (often characterized as the minimum detectable change or just noticeable difference \cite{azocar2017stiffness}), we restrict each parameter to belong to a discretized set with step sizes defined by $d_i \in \R$. This restriction also enables computational tractability.

Using this notation, each individual dimension of the entire space of possible parameter combinations is defined as:
\begin{align}
    \actions_i = \{a_i^{\text{min}} + n d_i \mid n \in \mathbb{N}_0 \textrm{ and } a_i^{\text{min}}+n d_i \leq a_i^{\text{max}}\},
\end{align}
for $i = \{1,\dots,\d\}$. We refer to the entire space of discrete parameter combinations as the \textit{action space} $\actions$ with with cardinality $|\actions| = \prod_{i = 1}^{\d} |\actions_i|$. Each unique vector of parameter values is referred to as an \textit{action}, denoted as $\act := [a_1, \dots, a_\d] \in \R^\d$.  
Lastly, we assume that there exists some unknown reward function $\util: \R^\d \to \R$ which maps each action $\act$ to a latent reward, i.e. the human's valuation of the action. The restriction of $r$ to $\actions$ is denoted as $\utilvec \in \R^{|\actions|}$. 

\subsection{Preference-Based Learning Architecture}
As illustrated in Fig. \ref{fig: intro-platforms}, human-in-the-loop preference-based learning refers to the iterative process of querying a human for qualitative feedback and utilizing this feedback to either execute actions that are likely to maximize the users latent reward function $\util$ (regret minimization), or to characterize the latent reward function (information gain). Three learning algorithms were introduced in prior work to individually accomplish the following: CoSpar for regret minimization \cite{tucker2020preference}; LineCoSpar for regret minimization in high-dimensional action spaces \cite{tucker2020human}; and ROIAL for information gain constrained to a \textit{region of interest} \cite{li2020roial}.


To provide a unified treatment of these previously disjoint learning algorithms, we structure the POLAR toolbox around the three main components that the algorithms share: 1) selecting new actions to sequentially execute on the system, 2) collecting user feedback corresponding to the sampled actions and 3) modeling the underlying reward function as a Gaussian Process (GP) using the provided feedback. These components are implemented in POLAR as the following classes:

\newclass{Sampling} This class selects new actions using one of three sampling techniques: 1) Thompson sampling for regret minimization; 2) information gain across a \textit{region of interest}; or 3) random sampling for comparison purposes only.

\newclass{UserFeedback} This class queries the human for up to three forms of feedback: 1) pairwise preferences; 2) user suggestions (coactive feedback); and 3) ordinal labels.

\newclass{GP} This class approximates the underlying reward function as a Gaussian Process using the provided user feedback.


\section{Modeling User Feedback}
\label{Sec: feedback}
We take advantage of three subjective feedback mechanisms: pairwise preferences, coactive suggestions, and ordinal labels. In this section, we formally define and compare how we model each of these mechanisms. 

\subsection{Preference Feedback}
\begin{figure}[tb]
    \centering
    \subfloat[]{\includegraphics[width=0.45\linewidth]{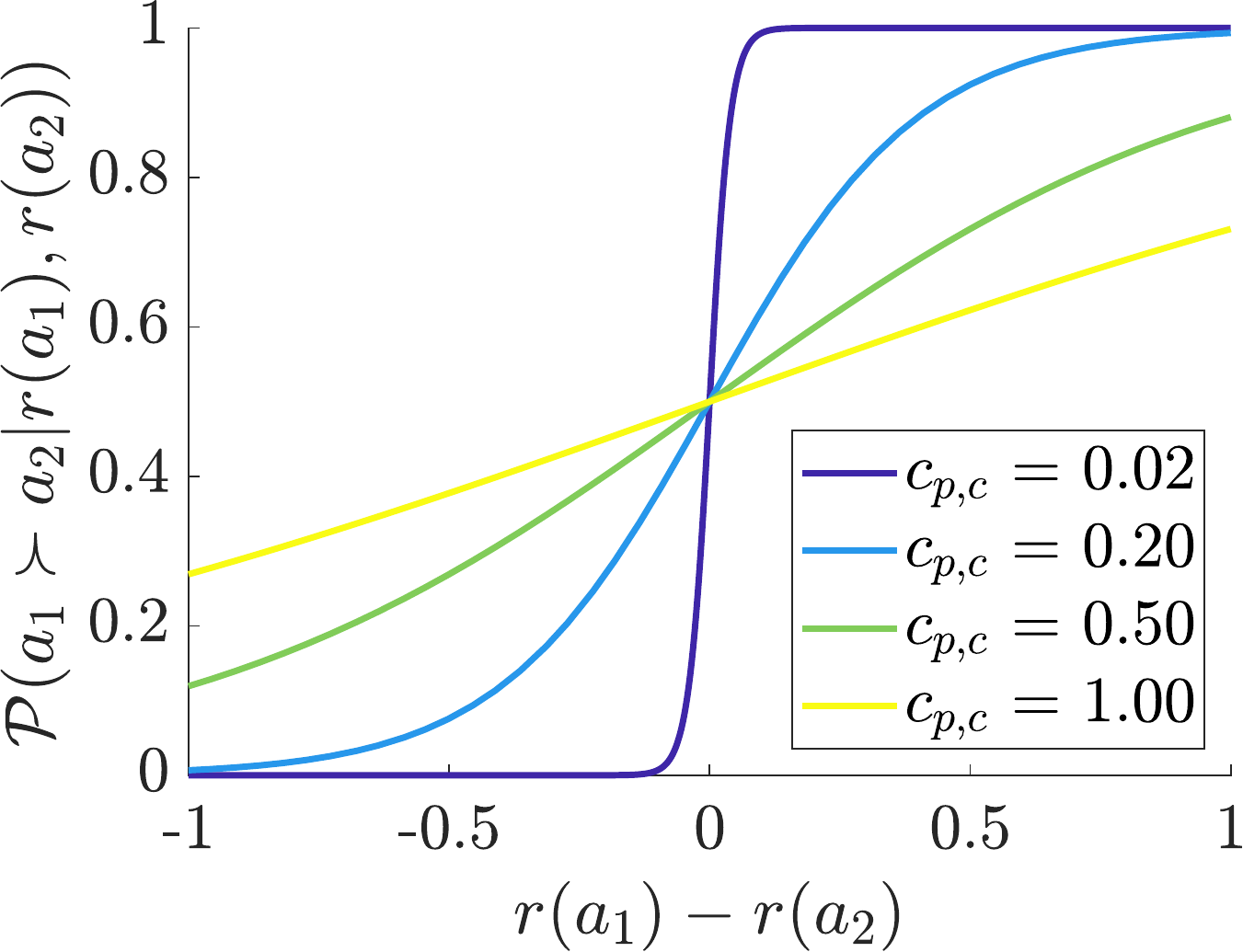}%
    \label{fig: preflikelihood}}
    \hfil
    \subfloat[]{\includegraphics[width=0.45\linewidth]{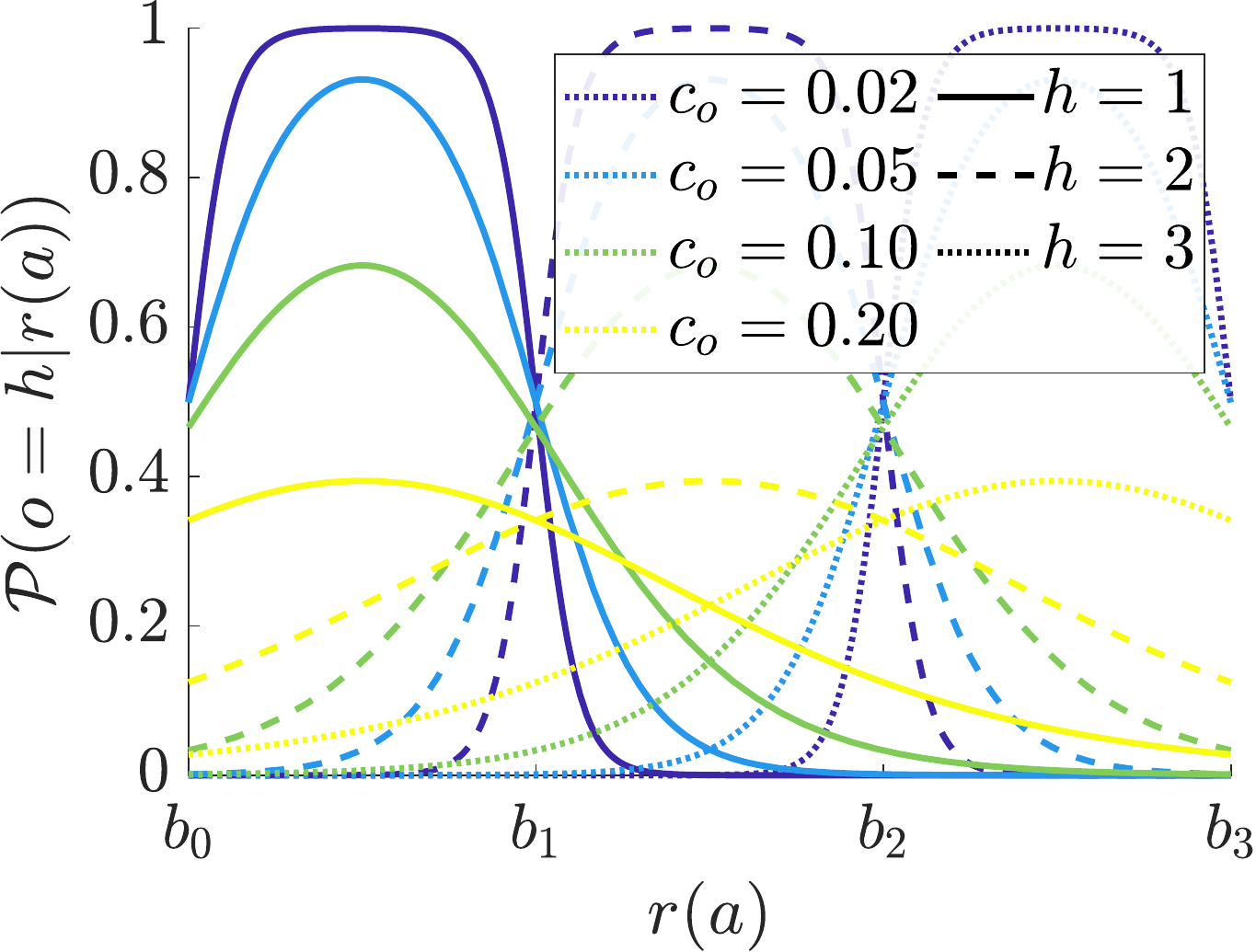}%
    \label{fig: ordlikelihood}}
    \caption{Illustration of the effect of the noise parameter on the likelihood function for a) preference and coactive feedback ($c_p$ and $c_c$), and b) ordinal feedback ($c_o$).}
    \label{fig: likelihoods}
\end{figure}

Preferences are defined as pairwise comparisons (i.e. ``Does the user prefer action a or action b''). A preference between two actions $\act_1$ and $\act_2$, is denoted as $p = \act_1 \succ \act_2$ if action $\act_1$ is preferred, or $p = \act_2 \succ \act_1$ if action $\act_2$ is preferred. The probability of a user giving a pairwise preference $\act_1 \succ \act_2$ given an underlying reward function $\util$ is modeled using the likelihood function \cite{chu2005preference}:   
\begin{align*}
    \P(\act_{1} \succ \act_{2} \mid \util(\act_{1}), \util(\act_{2})) = 
    \begin{cases} 
      1 & \text{if } \util(\act_{1}) \geq \util(\act_{2}), \\
      0 & \text{otherwise},
   \end{cases}
\end{align*}
which captures the preference relations given ideal noise-free preference feedback. However, user preferences are expected to be corrupted by noise. Thus, we model the preferences as being contaminated by noise through the likelihood function:
\begin{align*}
   &  \P(\act_{1} \succ \act_{2} \mid \util(\act_{1}), \util(\act_{2})) = g\left(\frac{\util(\act_{1})-\util(\act_{2})}{c_p}\right), 
\end{align*}
where $g(\cdot): \mathbb{R} \to (0,1)$ can be any monotonously-increasing activation function, and $c_p > 0$ quantifies noisiness in the preferences. The effect of $c_p$ had on the preference likelihood function is illustrated in Fig. \ref{fig: preflikelihood}. Two example link functions include the standard normal cumulative distribution function and the sigmoid function:
\begin{align*}
    g_{\text{Gaus}}(x) &:= \int_{-\inf}^{x} \mathcal{N}(\gamma; 0,1) d\gamma, \\
    g_{\text{sig}}(x) &:= \frac{1}{1+\exp(-x)}.
\end{align*}
We empirically found that using $g_{\text{sig}}(x)$ resulted in improved performance because of its heavier-tailed distribution. The preference likelihood function with $g_{\text{sig}}$ is illustrated in Fig. \ref{fig: preflikelihood} for preference noise parameters $c_p$. Note that the choice of $c_p$ depends both on the expected preference noise as well as the range of the underlying utility function $\util$.

A collection of pairwise preferences is denoted as:
\begin{align*}
    \prefdata &:= \{\act_{k1} \succ \act_{k2} \mid k = 1, \dots, K \},
\end{align*}
where $a_{k1}$ and $a_{k2}$ denote the two actions corresponding to the $k$\ts{th} pairwise preference in a set of $K$ total comparisons. The likelihood function for the entire set $\prefdata$ is then calculated as the product of each individual likelihood:
\begin{align*}
    \P(\prefdata \mid \util) &= \prod_{k=1}^K \P({\act}_{k1} \succ \act_{k2} \mid \util({\act}_{k1}), \util(\act_{k2})).
\end{align*}
This function is known as the preference likelihood function and is used later to approximate the posterior distribution.

\subsection{Coactive Feedback}
User suggestions, also known as coactive feedback, can be incorporated into the learning framework by treating user suggested improvements as implicit preferences. In this context, coactive feedback can be thought of as preference feedback between an action $\bar{\act}$ suggested by the user and the sampled action $\act$. The underlying assumption of coactive feedback is that $\util(\bar{\act}) > \util(\act)$. This approach resembles the \textit{coactive learning} framework \cite{shivaswamy2015coactive}, first proposed in  \cite{shivaswamy2012online}, in which the user identifies an improved action as feedback to each presented action. The combination of preference and coactive feedback is termed \textit{mixed-initiative} learning \cite{wolfman2001mixed, lester1999lifelike}. Coactive learning has been applied to robot trajectory planning \cite{jain2015learning, somers2016human}, but was only first, to our knowledge, applied with preference learning in \cite{tucker2020preference}. 

We denote a single user suggestion as $c = \bar{\act} \succ \act$. As with preferences, coactive feedback is modeled with the assumption that it is corrupted by noise:
\begin{align*}
    \P(\bar{\act} \succ \act \mid \util(\bar{\act}), \util(\act)) = g\left(\frac{\util(\bar{\act})-\util(\act)}{c_c}\right), 
\end{align*}
where $g(\cdot): \mathbb{R} \to (0,1)$ is the same link function as with preference feedback, and $c_c > 0$ quantifies noisiness in the coactive suggestions. The hyperparameter $c_c$ has an identical effect on the coactive likelihood function as $c_p$ had on the preference likelihood function, illustrated in Fig. \ref{fig: preflikelihood}. 

We recommend setting $c_c > c_p$ since in general, coactive feedback is more prone to noise compared to preference feedback. For example, in the case of exoskeleton user feedback, a user may provide a suggestion (i.e. ``I would prefer a faster gait'') that would no longer be true once they experienced the walking (i.e. ``Actually I feel uncomfortable with the faster gaits''). However, incorporating suggestions increases the amount of information obtained from the same number of sampled actions, thus increasing the sample-efficiency of the algorithm. 

A collection of user suggestions can either be stored with the preference dataset, or in their own dataset. For clarity, we will use the latter and define a set of user suggestions as:
\begin{align*}
    \coacdata := \{ \bar{\act}_{l} \succ \act_{l} \mid l = 1, ....., L\},
\end{align*}
where $\bar{\act}_{1}$ is the suggested action being compared to the sampled action $\act_{l}$, with this comparison being the $l$\ts{th} suggestion in a set of $L$ total suggested actions. As with user preferences, the coactive likelihood function is calculated as the product of each likelihood function for $L$ user suggestions:
\begin{align*}
    \P(\coacdata \mid \util) &= \prod_{l=1}^L \P(\bar{\act}_{l} \succ \act_{l} \mid \util(\bar{\act}_{l}), \util(\act_{l})).
\end{align*}

\subsection{Ordinal Feedback}
Ordinal feedback assigns an ordered label to each sampled action, with the ordinal labels corresponding to one of $\r$ ordinal categories. Thus, ordinal labels partition the set of all possible actions into $\r$ sets denoted as $\ord_i$ for $i = 1, \dots, \r$. We define the boundaries between ordinal categories using thresholds $-\infty = b_0 < b_1 < \dots < b_{\r} = \infty$. Thus, in the case of ideal noise-less feedback, ordinal labels are determined by the likelihood function:
\begin{align*}
    \P \big( (\act,o) \mid \util(a) \big) = 
    \begin{cases}
    1 & \text{if } b_{o-1} \leq \util(\act) < b_{o} \\
    0 & \text{otherwise}
    \end{cases},
\end{align*}
where we denote $(\act,o)$ as the ordinal label $o \in \{1,\dots,\r\} \subset \R$ provided for the sampled action $\act$. As with preference and coactive feedback, we modify this simplified likelihood function to account for noise as in \cite{chu2005gaussian}:
\begin{align*}
    \P\big( (\act,o)  \mid \util(\act) \big) =g\left( \frac{b_{o} - \util(\act)}{c_o} \right)  - g \left( \frac{b_{o-1} - \util(\act)}{c_o}\right),
\end{align*}
with $g(\cdot): \mathbb{R} \to (0,1)$ a link function as before, and $c_o > 0$ quantifying noisiness in the ordinal labels. The effect of $c_o$ on the likelihood function is illustrated in Fig. \ref{fig: ordlikelihood}. Since previous work has shown numerical scores (such as ordinal labels) to be less reliable than pairwise preferences for subjective human feedback, we set $c_o > c_c > c_p$ \cite{sui2018advancements,joachims2005accurately}. 

We define a dataset of $M$ ordinal labels as
\begin{align*}
    \orddata := \{ (\act_m,o_m)\mid m = 1, ....., M\},
\end{align*}
with each ordinal label $o_m$ given for the corresponding action $\act_m$. As with the preference and coactive likelihood functions, the ordinal likelihood function is calculated as the product of the $M$ individual likelihoods:
\begin{align*}
    \P(\orddata \mid \util) &= \prod_{m=1}^M \P\big( (\act_m,o_m) \mid \util(\act_m) \big).
\end{align*}

\section{Estimating the Underlying Utility Function}
\label{Sec: modeling}

The second component of the learning framework is to use the obtained user feedback to estimate the vectorized underlying utility function, $\utilvec \in \R^{|\actions|}$, as $\hat{\utilvec} \in \R^{|\actions|}$. 
We compute $\hat{\utilvec}$, as the maximum a posteriori (MAP) estimate:
\begin{align*}
    \hat{\utilvec} = \bm{\util}_{\text{MAP}} := \argmax_{\utilvec \in \R^{|\actions|}} \P(\utilvec \mid \data),
\end{align*}
where $\P(\utilvec \mid \data)$ is the posterior distribution of $\utilvec$ given the collection of user feedback $\data = \prefdata \cup \coacdata \cup \orddata$. We choose to approximate $\hat{\utilvec}$ as a Gaussian process because it enables $\utilvec$ to be modeled as a Bayesian posterior over a class of smooth, non-parametric functions. We approximate $\P(\utilvec \mid \data)$ as the Gaussian distribution $\N(\mu,\Sigma)$, which is derived from the preference-based Gaussian process model of \cite{chu2005preference}. This allows us to approximate $\hat{\utilvec} = \mu$. 

In this section, we will first discuss how we model the Bayesian posterior $\P(\utilvec \mid \data)$, followed by how we approximate this distribution as Gaussian. The end result is an optimization problem which can be computationally expensive depending on the number of actions in $\actions$ over which $\utilvec \in \R^{|\actions|}$ inferred. Thus, to remain tractable in high-dimensional action spaces, we only infer the reward function over a subset of actions, denoted $\utilvec_{\sub} \in \R^{|\sub|}$ for some $\sub \subset \actions$. We will discuss this restriction at the end of this section.

\subsection{Modeling the posterior probability}
By assuming conditional independence of the feedback mechanisms, we can use Bayes rule to model the posterior of the utilities $\utilvec$ as proportional to the product of the individual likelihood terms and the Gaussian prior:
\begin{align*}
    \P(\utilvec | \data) \propto \P(\prefdata | \utilvec) \P(\coacdata | \utilvec) \P(\orddata | \utilvec) \P(\utilvec),
\end{align*}
with each likelihood calculated as in Sec. \ref{Sec: feedback}. We define a Gaussian prior over $\utilvec$: 
\begin{align*}
    \P(\utilvec) = \frac{1}{(2\pi)^{\frac{|\actions|}{2}} | \Sigma^{\text{pr}}|^{\frac{1}{2}}} \exp \left( -\frac{1}{2}\utilvec^{\top} (\Sigma^{\text{pr}})^{-1} \utilvec \right),
\end{align*}
where $\Sigma^{\text{pr}} \in \R^{|\actions| \times |\actions|}$ is the prior covariance matrix with $[\Sigma^{\text{pr}}]_{ij} = \mathcal{K}(\act_i,\act_j)$ and $\mathcal{K}$ being a kernel of choice. In our work, we select $\mathcal{K}$ to be the squared exponential kernel,
\begin{align*}
    \mathcal{K}_{\text{SE}}(\act,\act') = \sigma^2 \exp\left( - \frac{(\act - \act')^2}{2l^2} \right),
\end{align*} 
where $\sigma \in \R$ is the output variance hyperparameter and $l \in \R^\d$ is a vector of lengthscales for each dimension of $\actions$. The output variance $\sigma$ dictates the expected average distance the underlying function is away from its mean. The lengthscales $l_i$ for $i = 1,\dots,\d$ are hyperparameters that dictate the expected ``wiggliness'' of the underlying function in each dimension.

\subsection{Approximating the posterior distribution as Gaussian}

Since we leverage more types of feedback than just pairwise preferences, we extend the method introduced in \cite{chu2005preference} towards approximating $\P(\utilvec \mid \data)$. This extension redefines the Gaussian distribution $\N(\mu,\Sigma)$ as:
\begin{align*}
    \mu &= \bm{\util}_{\text{MAP}} := \argmax_{\utilvec} \P(\utilvec \mid \data), \\
    \Sigma &= ((\prior)^{-1} + \Lambda_{\text{MAP}})^{-1},
\end{align*}
where $\Lambda_{\text{MAP}}$ is the matrix $\Lambda \in \R^{|\actions| \times |\actions|}$ evaluated at $\bm{\util}_{\text{MAP}}$, with the composition of $\Lambda$ extended to: 
\begin{align*}
    [\Lambda]_{ij} := & \sum_{k=1}^K\frac{-\partial^2 \ln \P \left({\act}_{k1} \succ \act_{k2} \mid \utilvec({\act}_{k1}), \utilvec(\act_{k2}) \right) }{\partial \utilvec(\act_i)\utilvec(\act_j)} \\
    & + \sum_{l=1}^L\frac{-\partial^2  \ln \P(\bar{\act}_{l} \succ \act_{l} \mid \utilvec(\bar{\act}_{l}), \utilvec(\act_{l}))}{\partial \utilvec(\act_i)\utilvec(\act_j)} \\
    & + \sum_{m=1}^M\frac{-\partial^2  \ln \P((\act_m,o_m) \mid \utilvec(\act_m))}{\partial \utilvec(\act_i)\utilvec(\act_j)}.
\end{align*}
Here, $[\Lambda]_{ij}$ denotes the $ij$\ts{th} element of $\Lambda$ for $i = 1, \dots, |\actions|$ and $j = 1, \dots, |\actions|$. 

\subsection{Solving for the MAP estimate using convex programming}
Approximating $\P(\utilvec \mid \data)$ as a Gaussian distribution centered on $\bm{\util}_{\text{MAP}}$ with the covariance matrix $\Sigma$ is equivalent to the Laplace approximation of the functional:
\begin{align*}
    \S(\utilvec) = &-\sum_{k=1}^K \ln \P({\act}_{k1} \succ \act_{k2} \mid \utilvec({\act}_{k1}), \utilvec(\act_{k2}))  \\
    &- \sum_{l=1}^L \ln \P(\bar{\act}_{l} \succ \act_{l} \mid \utilvec(\bar{\act}_{l}), \utilvec(\act_{l}))  \\
    & -\sum_{m=1}^M \ln \P((\act_m,o_m) \mid \utilvec(a_m))
    +\frac{1}{2}\utilvec^{\top} (\Sigma^{\text{pr}})^{-1} \utilvec.
\end{align*}
The MAP estimate $\bm{\util}_{\text{MAP}}$ is computed as the minimizer of $\S(\utilvec)$ using a convex program. The details of this convex program are provided in the appendix.

\subsection{Restricting the posterior distribution to a subset $\sub$} 
Solving for $\hat{\utilvec}$ is computationally-expensive and can even be intractable for high-dimensional action spaces. Moreover, we cannot leverage existing work with high-dimensional Gaussian process learning since it requires quantitative feedback \cite{kandasamy2015high, wang2013bayesian}. Thus, to maintain computational tractability in high-dimensional action spaces, we simply restrict the posterior to only a subset of the discrete action space $\actions$, denoted as the set $\sub \subset \actions$. The restricted utility function is denoted $\utilvec_{\sub}\in \R^{|\sub|}$. 

Since the process of computing $\hat{\utilvec}_{\sub}$ is nearly identical to computing $\hat{\utilvec}_{\actions}$, the details are omitted here, but can be found in the appendix. Note that the specific composition of the subset $\S$ depends on the sampling strategy of the framework, and will thus be discussed in the next section.

\section{Selecting New Actions}
\label{Sec: sampling}
Since human-in-the-loop frameworks only collect user feedback for sampled actions, the process of selecting these actions is critical to the learning performance. In this section we will outline each of the two sampling techniques used in the POLAR toolbox: Thompson sampling and information gain across a region of interest.

\subsection{Sampling method for regret minimization}
There are many existing sampling methods aimed at regret minimization, including Thompson sampling \cite{thompson1933likelihood} and upper confidence bound (UCB) algorithms \cite{lai1985asymptotically}. Existing work has explored extending UCB algorithms for settings with relative feedback (such as pairwise preferences), aptly named Relative Upper Confidence Bound (RUCB) \cite{zoghi2014relative}. However, in our work we utilize Thompson Sampling since it slightly favors exploitation compared to RUCB in the presence of relative feedback \cite{tucker2020preference}. 

The general concept of Thompson sampling is to determine $n$ actions to query by drawing $n$ samples from a given distribution. Specifically, the probability of selecting action $\act_i$ corresponds with it's probability of being optimal. Thus, as the uncertainty of the distribution shrinks, the action maximizing the mean of the distribution has a higher probability of being selected. 


POLAR conducts Thompson sampling using the following procedure 
(assuming dimensionality reduction). In each iteration $i$, $n$ samples are drawn from the distribution:
\begin{align*}
    \utilvec^k_i \sim \N(\mu_\sub, \Sigma_\sub) \quad \forall k = 1,\dots, n.
\end{align*}
Then, the sampled actions $\{\act^1_i, \dots, \act^n_i\}$ are selected as the actions maximizing the drawn samples:
\begin{align*}
    \act^k_i = \argmax_{\act \in \sub} \utilvec^k_i(\act) \quad \forall k = 1,\dots, n .
\end{align*}

While the restriction of the posterior to a subset $\sub \subset \actions$ addresses the issue of computational tractability, it is critical to construct $\sub$ in a way that preserves regret minimization. Inspired from \cite{kirschner2019adaptive}, we construct $\sub := \randline \cup \vis$ as the union of a one-dimensional subspace $\randline \subset \actions$ with all previously sampled actions $\vis$. Importantly, $\randline$ is generated such that it intersects the most recent estimate of the optimal action ($\act^* = \argmax_{\act \in \sub} \mu_{\sub}$). While this dimensionality reduction technique limits the set of possible sampled actions to those within $\sub$, \cite{kirschner2019adaptive} found that it yields state-of-the-art performance with significantly reduced computational complexity.

\subsection{Sampling method for preference characterization}
While there are many existing sampling approaches aimed at characterizing the latent reward function, we utilize a custom sampling approach \cite{li2020roial} in order to: 1) prioritize sample efficiency; 2) utilize subjective feedback; 3) avoid low-utility actions; and 4) select actions that lead to easy to answer questions and reliable feedback. This custom acquisition function resembles information gain but limited to actions within a ``Region of Interest'' (ROI).

Information gain selects actions that maximize the mutual information between the underlying utility function and the users feedback, thus learning the underlying utility function as efficiently as possible. It has been shown that selecting an entire sequence of actions to optimize this mutual information is NP-hard \cite{ailon2012active}. However, previous work has shown that state-of-the-art performance can be achieved via a greedy approach which only optimizes one action to compare with already sampled past actions. Thus, in our work we also only select one action at a time using information gain which is compared with past actions. It is possible to extend this method to more than one action sampled in each iteration but the problem quickly becomes intractable.

POLAR conducts information gain using the following procedure. In iteration $i$, $n$ new actions are selected as the solutions to the maximization problem:
\begin{align*}
    \act_i^k = \argmax_{\act \in \sub_{\text{ROI}}} I (\utilvec; o_i^k, \bm{p}_i^k | \prefdata, \orddata, \act), \quad \forall k = 1, \dots, n,
\end{align*}
where $o_i^k$ is the ordinal label associated with the sampled action $\act^k_i$, and $\bm{p}_i^k = \{p_i^1, \dots, p_i^{n+b-1}\}$ is the set of pairwise preferences between $\act_i^k$ and each action within the set $\{\{\{\act_i^1, \dots, \act_i^n\} \setminus \act_i^k\} \cup \act_{\text{buffer}}\}$, where $\act_{\text{buffer}}$ represents the set of $b$ previously sampled actions stored in a ``buffer''. This set of buffer actions can be interpreted as the previously sampled actions that the user can reliably remember. 

This maximization problem can be equivalently written in terms of information entropy:
\begin{align*}
    \act_i^k = &\argmax_{\act \in \sub_{\text{ROI}}} ~H(o^k_i,\bm{p}_i^k | \prefdata, \orddata, \act) \\
    & - \E_{\utilvec | \prefdata \orddata} [H(o_i^k,\bm{p}_i^k | \prefdata,\orddata, \act ,\utilvec)], \quad \forall k = 1,\dots,n.
\end{align*} 
In this expression, the first term can be interpreted as the model's uncertainty about a given action's ordinal label and preferences. The sampled action is aimed at maximizing this term because an action with high uncertainty regarding its feedback yields potentially valuable information regarding the underlying utility function. The second term can be interpreted as the user's expected uncertainty regarding the feedback. Therefore, the sampled action is aimed at yielding pairwise comparisons that are easy for the user to provide. By combining both terms, the sampled actions result in queries that are both informative and easy for users. 

\newsubsec{Limiting exploration to a region of interest}
The region of interest (ROI) is defined as the set of actions $\sub_{\text{ROI}} \in \sub \subseteq \actions$, restricted to the set $\sub$ over which the posterior $\N(\mu_\sub,\Sigma_\sub)$ is constructed, that satisfy the criteria:
\begin{align*}
    \mu_\sub(\act) + \lambda \Sigma_\sub(\act) > b_{\text{ROI}}.
\end{align*}
The constant $b_{\text{ROI}} \in \R$ is the user-selected threshold that separates the ordinal categories belonging to the ROA and the ordinal categories belonging to the complement of the ROA. We term this complement the region of interest (ROI). 
The user also defines $\lambda$, a hyperparameter that determines the algorithm's conservatism in estimating the ROI. Smaller values of $\lambda$ lead to more conservative estimates of the ROA. 

\section{Demonstration of POLAR in Simulation}
\label{Sec: simulation}
Another advantage of the POLAR toolbox is that can provide synthetic feedback, dictated by a user-provided synthetic objective function, to enable simulations of the learning framework. These simulations can provide insight into the effect of various various settings and hyperparameters on the overall learning performance. For example, we will present simulations that compare the sampling methods and included feedback types. To facilitate with running these comparisons, the POLAR toolbox includes the following classes:

\newclass{SyntheticFeedback} This class returns synthetic feedback based on a pre-specified \textit{true} underlying utility function. The noisiness of the synthetic feedback can be adjusted through simulated noise parameters.

\newclass{Compare} This class runs several instances of preference-based learning using synthetic feedback to compare specific settings of framework. 

\newclass{EvaluatePBL} This class computes various metrics used to evaluate the performance of the simulated learning framework including error of the predicted optimal action compared to the true optimal action, instantaneous regret, and computation time of the posterior update.

\begin{figure}[tb]
    \centering
    \includegraphics[width=0.95\linewidth]{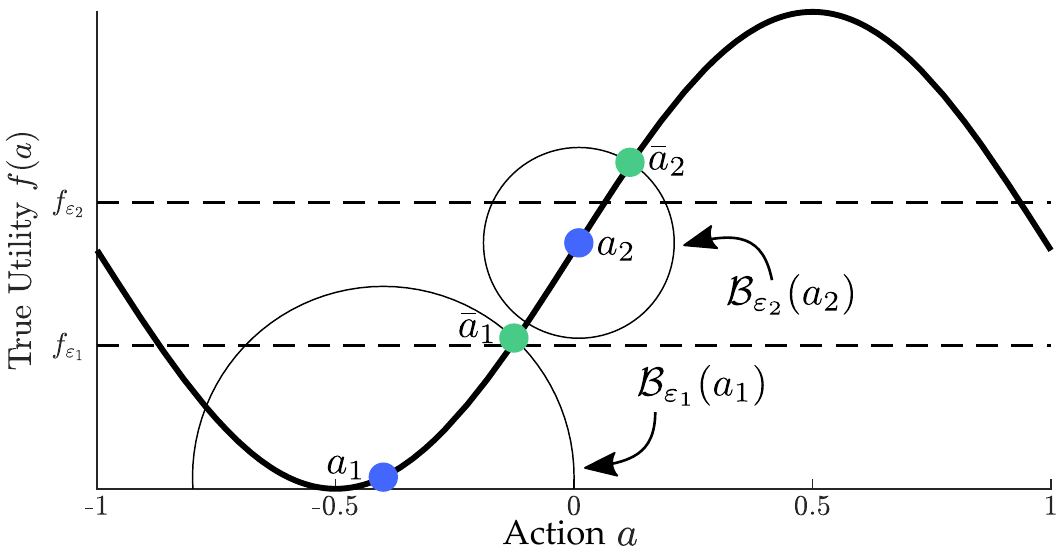}
    \caption{Simple illustration of synthetically generated coactive feedback where $\act_1$ and $\act_2$ are two sampled actions with respective coactive actions $\bar{\act}_1$ and $\bar{\act}_2$. The coactive actions are selected as the actions with maximum utility within $\mathcal{B}_{\epsilon}(\act)$, with $\epsilon = \epsilon_1$ for $\act \leq f_{\epsilon_1}$ and $\epsilon = \epsilon_2$ for $f_{\epsilon_1} < \act \leq f_{\epsilon_1}$. Here, $\epsilon_1,\epsilon_2,f_{\epsilon_1},f_{\epsilon_2} \in \R$ are all hyperparameters.} 
    \label{fig: synthcoac}
    \vspace{-2mm}
\end{figure}

\begin{figure*}[t]
    \centering
    \includegraphics[width=\linewidth]{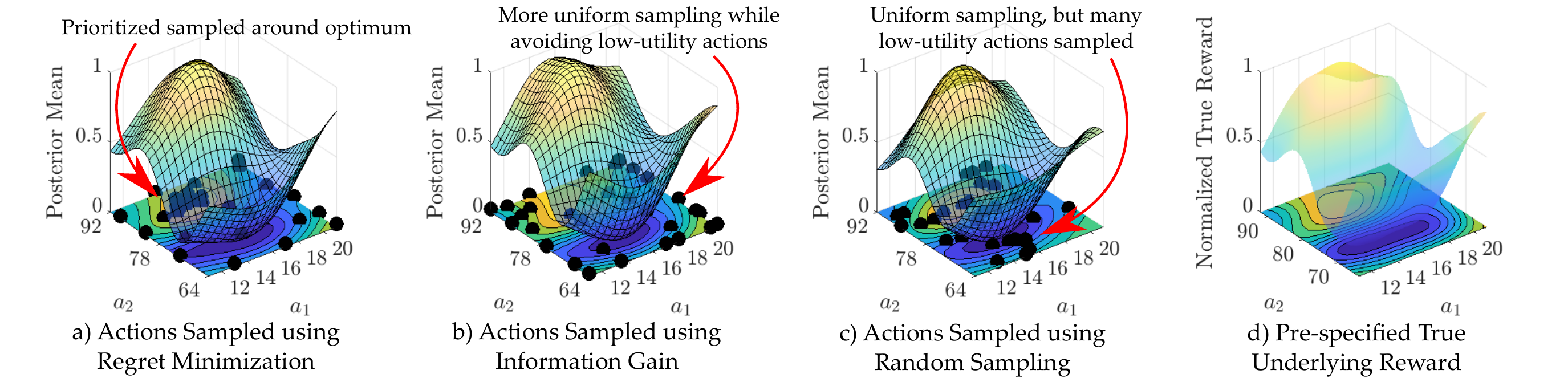}
    \caption{Simulations were run to compare the three sampling strategies: a) regret minimization, b) information gain, and c) uniform random sampling; with d) illustrating the true underlying reward function. As shown, the regret minimization sampling strategy (a) resulted in more high-utility actions being sampled, while the information gain sampling strategy (b) resulted in a more uniform sampling of actions while avoiding low-utility actions.}
    \label{fig:samplingcomparison}
    \vspace{-4mm}
\end{figure*}

\subsection{Generating Synthetic Feedback}
Synthetic feedback is generated in place of human-provided feedback using the POLAR class \texttt{SyntheticFeedback}. This synthetic feedback is generated with respect to a  user-provided \textit{true} utility function $\true: \mathbb{R}^\d \to \R$. In this section, we will explain how each of the three feedback types are synthetically generated.

\newsubsec{Synthetic pairwise preferences} Synthetic preference feedback is generated using the preference likelihood function,
\begin{align*}
    \prefdata^{(i)} = \begin{cases}
    \act_{1} \succ \act_{2}  & w.p. \quad g\left(\frac{\true(\act_1)-\true(\act_2)}{\tilde{c}_p}\right) \\
    \act_{2} \succ \act_{1}  & w.p. \quad 1 - g\left(\frac{\true(\act_1)-\true(\act_2)}{\tilde{c}_p}\right)
    \end{cases} 
\end{align*}
for any two actions $\act_{1}$ and $\act_{2}$ being compared during iteration $i$, a parameter $\tilde{c}_p$ on the level of noise in the synthetic preference feedback, and any link function $g(\cdot): \mathbb{R} \to (0,1)$. Note that we designate $\tilde{c}_p$ to be the noise of the synthetically generated preferences, while $c_p$ is the noise hyperparameter used by the algorithm.

\newsubsec{Synthetic user suggestions} Synthetic coactive feedback is generated as a suggested action $\bar{\act}_c$ with respect to a sampled action $\act_c$ that satisfies the condition $\true(\bar{\act}_c) > \true(\act_c)$. There are many ways that such coactive actions $\bar{\act}_c$ can be synthetically generated. In our work, we generated synthetic feedback using the following criteria, as illustrated in Fig. \ref{fig: synthcoac}, 
\begin{align*}
    \bar{a}_c = \begin{cases}
          \bar{\act}_c = \argmax_{\act \in \mathcal{B}_{\epsilon_1}(\act_c)} \true(\act) & \text{if } \true(\act_c) \leq \true_{\epsilon_1}, \\
          \bar{\act}_c = \argmax_{\act \in \mathcal{B}_{\epsilon_2}(\act_c)} \true(\act) & \text{if } \true_{\epsilon_1} \leq \true(\act_c) \leq \true_{\epsilon_2}, \\
          \bar{\act}_c = \emptyset & \text{otherwise}.
      \end{cases}
\end{align*}
Here, $\mathcal{B}_{\epsilon_1}(\act_c)$, and $\mathcal{B}_{\epsilon_2}(\act_c)$ are Euclidean balls of respective radii $\epsilon_1$ and $\epsilon_2$ centered at $\act_c$. The epsilon values $\epsilon_1$ and $\epsilon_2$ are predetermined hyperparameters that dictate how far around the sampled action the human can predict in terms of the underlying utilities. 
Similarly, $\true_{\epsilon_1}$ and $\true_{\epsilon_2}$ are predetermined thresholds that dictate when the respective coactive feedback is provided.  The justification behind these thresholds is that a user is more likely to provide coactive feedback when the relative value of $\true(\act_c)$ is low.

Noisy coactive feedback is modeled as with preferences:
\begin{align*}
    \coacdata^{(i)} = \begin{cases}
    \bar{\act}_{c} \succ \act_{c}  & w.p. \quad g\left(\frac{\true(\bar{\act}_c)-\true(\act_c)}{\tilde{c}_c}\right), \\
    \act_{c} \succ \bar{\act}_{c}  & w.p. \quad 1 - g\left(\frac{\true(\bar{\act}_c)-\true(\act_c)}{\tilde{c}_c}\right).
    \end{cases} 
\end{align*}

\newsubsec{Synthetic ordinal labels} Lastly, synthetic ordinal labels are generated using the ordinal likelihood function with a parameter for simulated noise $\tilde{c}_o$:
\begin{align*}
    \orddata^{(i)} = \begin{cases}
    (\act_m,o_m) & w.p. \quad g\left( \frac{\tilde{b}_{o_m} - \true(\act_m)}{\tilde{c}_o} \right) \\
    &  \hspace{10mm} - g \left( \frac{\tilde{b}_{o_m-1} - \true(\act_m)}{\tilde{c}_o}\right),
    \end{cases} 
\end{align*}
for any action $\act_m$ being sampled during iteration $i$, with $o_m$ being one of $\r$ ordinal labels $o_m \in \{1, \dots, \r\}$, and $-\infty < \tilde{b}_0, \dots, \tilde{b}_{\r} < \infty$ being the ordinal thresholds of the true underlying ordinal categories.

\subsection{Comparing regret minimization and information gain}
Fig. \ref{fig:samplingcomparison} illustrates the difference between the learning algorithms performance when utilizing regret minimization compared to information gain. As shown, the regret minimization sampling strategy (Fig. \ref{fig:samplingcomparison}a) results in more high-utility actions being sampled, while the information gain sampling strategy (Fig. \ref{fig:samplingcomparison}c) results in a more uniform sampling of actions while avoiding low-utility actions. The results of uniform random sampling is also illustrated for comparison purposes in the third column of Fig. \ref{fig:samplingcomparison}. 

\subsection{Simulation results for various combinations of feedback}
\begin{figure}[!t]
\centering
\includegraphics[width=\linewidth]{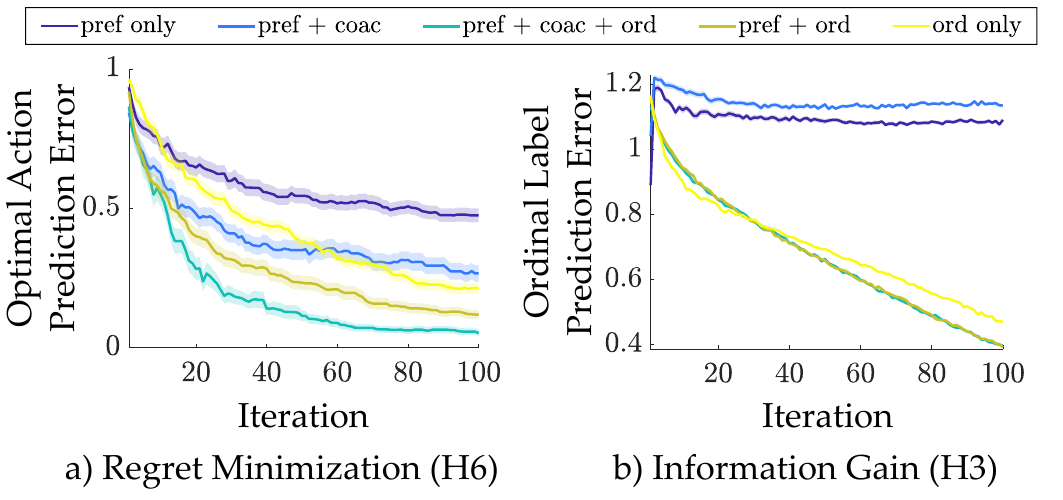}
\caption{The figures illustrate the effect of included feedback types towards a) regret minimization for the Hartmann-6 synthetic function, and b) information gain for the Hartmann-3 synthetic function. The plots show the mean and standard error of the results, with 50 simulations per condition.}
\label{fig: sim_feedback}
\vspace{-4mm}
\end{figure}

Next we present the simulation results comparing various combinations of provided feedback (i.e. only pairwise preferences compared to pairwise preferences and ordinal labels). Two sets of simulation were run: one which sampled new actions using regret minimization, and one using information gain. These sets respectively utilized the Hartmann 6-dimensional (H6) and Hartmann 3-dimensional synthetic function as the true objective function.
The learning performance of the regret minimization simulations is evaluated using the error between the reward of the true optimal action and the reward of the action maximizing the posterior mean (termed optimal action error). Alternatively, the learning performance of the information gain simulations is evaluated using the average error between the predicted ordinal labels and the true ordinal labels (termed ordinal label prediction error). As shown in Fig. \ref{fig: sim_feedback}, the learning performance for both regret minimization and information gain improves with each additional feedback type utilized.

\section{Example Applications}
\label{Sec: examples}
To emphasize the various uses of the POLAR toolbox, we next highlight previous experiments that utilized preliminary versions of the POLAR toolbox. These experiments deployed human-in-the-loop preference-based learning towards a wide range of robotic applications, as illustrated in Fig. \ref{fig: intro-platforms}.

\subsection{Tuning Parameters of Control Barrier Functions for Safe and Performant Quadrupedal Locomotion}
Often, there is a trade-off between conservative uncertainty bounds of a safety-critical controller and the performance of the resulting robotic behavior in the real world. However, this trade-off can be difficult to numerically optimize since real-world performance is subjective. Thus, human-in-the-loop preference-based learning can be useful tool for identifying safety-critical control parameters that result in human-preferred real world behavior, while still preserving safety guarantees. To demonstrate this, we deployed \textit{safety-aware} preference-based learning on the Unitree A1 Quadruped to tune parameters of a control barrier function \cite{cosner2022safety}.

\subsection{Optimizing and Characterizing User Comfort during Exoskeleton Locomotion}
Exoskeleton locomotion is challenging since it must be customized for each exoskeleton user. Furthermore, the relationship between exoskeleton user comfort and parameters of the walking gaits is poorly understood. Thus, preference-based learning is a sample-efficient technique for optimizing exoskeleton user comfort, as demonstrated using the Atalante exoskeleton \cite{tucker2020preference,tucker2020human}. Additionally, by conducting additional experiments aimed at information gain, we obtained the underlying preference landscapes of exoskeleton users to better understand the metrics underlying user comfort \cite{li2020roial}.

\subsection{Modifying Optimization Constraints to Realize Stable, Efficient, and Natural Bipedal Walking}
Traditionally, generating reference trajectories for bipedal robots requires a labor-intensive hand-tuning process in which a human operator must make assumptions about the relationship between the various parameters of the gait generation framework and the resulting experimental locomotion. However, it is easy for a human to visualize two instances of robotic walking and provide subjective feedback about which is \textit{better}. Thus, we also deployed preference-based learning to systematically identify parameters of a gait generation framework which led to stable, efficient, and natural bipedal locomotion on both the AMBER-3M bipedal robot \cite{tucker2021hzd} and the AMPRO3 dual-actuated prosthesis \cite{li2022natural}.

\subsection{Tuning Controller Gains for Bipedal Walking}
Similar to tuning optimization constraints, tuning controller gains also requires a laborious hand-tuning process and requires expert knowledge of how the various control parameters influence the resulting robotic behavior. Thus, we also demonstrated preference-based learning towards systematically identifying control parameters for stable and robust walking on the AMBER-3M and CASSIE bipeds \cite{csomay2021learning}.

\section{Conclusion}
\label{Sec: conclusions}

In this work, we introduce the unified open-source MATLAB toolbox POLAR (Preference Optimization and Learning Algorithms for Robotics) for sample-efficient human-in-the-loop preference-based learning. We present the details of learning framework, demonstrate its performance in simulation, and provide various example applications.


\bibliographystyle{IEEEtran}
\bibliography{references}

\begin{thebibliography}{10}
\providecommand{\url}[1]{#1}
\csname url@samestyle\endcsname
\providecommand{\newblock}{\relax}
\providecommand{\bibinfo}[2]{#2}
\providecommand{\BIBentrySTDinterwordspacing}{\spaceskip=0pt\relax}
\providecommand{\BIBentryALTinterwordstretchfactor}{4}
\providecommand{\BIBentryALTinterwordspacing}{\spaceskip=\fontdimen2\font plus
\BIBentryALTinterwordstretchfactor\fontdimen3\font minus
  \fontdimen4\font\relax}
\providecommand{\BIBforeignlanguage}[2]{{%
\expandafter\ifx\csname l@#1\endcsname\relax
\typeout{** WARNING: IEEEtran.bst: No hyphenation pattern has been}%
\typeout{** loaded for the language `#1'. Using the pattern for}%
\typeout{** the default language instead.}%
\else
\language=\csname l@#1\endcsname
\fi
#2}}
\providecommand{\BIBdecl}{\relax}
\BIBdecl

\bibitem{csomay2021learning}
N.~Csomay-Shanklin, M.~Tucker, M.~Dai, J.~Reher, and A.~D. Ames, ``Learning
  controller gains on bipedal walking robots via user preferences,''
  \emph{arXiv preprint arXiv:2102.13201}, 2021.

\bibitem{tucker2021hzd}
M.~Tucker, N.~Csomay-Shanklin, W.-L. Ma, and A.~D. Ames, ``Preference-based
  learning for user-guided hzd gait generation on bipedal walking robots,'' in
  \emph{2021 IEEE International Conference on Robotics and Automation
  (ICRA)}.\hskip 1em plus 0.5em minus 0.4em\relax IEEE, 2021.

\bibitem{tucker2020preference}
M.~Tucker, E.~Novoseller, C.~Kann, Y.~Sui, Y.~Yue, J.~W. Burdick, and A.~D.
  Ames, ``Preference-based learning for exoskeleton gait optimization,'' in
  \emph{2020 IEEE International Conference on Robotics and Automation
  (ICRA)}.\hskip 1em plus 0.5em minus 0.4em\relax IEEE, 2020, pp. 2351--2357.

\bibitem{ingraham2022role}
K.~A. Ingraham, C.~D. Remy, and E.~J. Rouse, ``The role of user preference in
  the customized control of robotic exoskeletons,'' \emph{Science robotics},
  vol.~7, no.~64, p. eabj3487, 2022.

\bibitem{tucker2020human}
M.~Tucker, M.~Cheng, E.~Novoseller, R.~Cheng, Y.~Yue, J.~W. Burdick, and A.~D.
  Ames, ``Human preference-based learning for high-dimensional optimization of
  exoskeleton walking gaits,'' \emph{arXiv preprint arXiv:2003.06495}, 2020.

\bibitem{li2020roial}
K.~Li, M.~Tucker, E.~B{\i}y{\i}k, E.~Novoseller, J.~W. Burdick, Y.~Sui,
  D.~Sadigh, Y.~Yue, and A.~D. Ames, ``Roial: Region of interest active
  learning for characterizing exoskeleton gait preference landscapes,''
  \emph{arXiv preprint arXiv:2011.04812}, 2020.

\bibitem{biyik2021aprel}
E.~B{\i}y{\i}k, A.~Talati, and D.~Sadigh, ``Aprel: A library for active
  preference-based reward learning algorithms,'' \emph{arXiv preprint
  arXiv:2108.07259}, 2021.

\bibitem{azocar2017stiffness}
A.~F. Azocar and E.~J. Rouse, ``Stiffness perception during active ankle and
  knee movement,'' \emph{IEEE Transactions on Biomedical Engineering}, vol.~64,
  no.~12, pp. 2949--2956, 2017.

\bibitem{chu2005preference}
W.~Chu and Z.~Ghahramani, ``Preference learning with {G}aussian processes,'' in
  \emph{Proceedings of the 22nd international conference on Machine learning},
  2005, pp. 137--144.

\bibitem{shivaswamy2015coactive}
P.~Shivaswamy and T.~Joachims, ``Coactive learning,'' \emph{Journal of
  Artificial Intelligence Research}, vol.~53, pp. 1--40, 2015.

\bibitem{shivaswamy2012online}
------, ``Online structured prediction via coactive learning,'' in
  \emph{Proceedings of the 29th International Conference on Machine
  Learning}.\hskip 1em plus 0.5em minus 0.4em\relax Omnipress, 2012, pp.
  59--66.

\bibitem{wolfman2001mixed}
S.~A. Wolfman, T.~Lau, P.~Domingos, P.~Domingos, and D.~S. Weld, ``Mixed
  initiative interfaces for learning tasks: {SMART}edit talks back,'' in
  \emph{Proceedings of the 6th International Conference on Intelligent User
  Interfaces}.\hskip 1em plus 0.5em minus 0.4em\relax ACM, 2001, pp. 167--174.

\bibitem{lester1999lifelike}
J.~C. Lester, B.~A. Stone, and G.~D. Stelling, ``Lifelike pedagogical agents
  for mixed-initiative problem solving in constructivist learning
  environments,'' \emph{User Modeling and User-Adapted Interaction}, vol.~9,
  no. 1-2, pp. 1--44, 1999.

\bibitem{jain2015learning}
A.~Jain, S.~Sharma, T.~Joachims, and A.~Saxena, ``Learning preferences for
  manipulation tasks from online coactive feedback,'' \emph{The International
  Journal of Robotics Research}, vol.~34, no.~10, pp. 1296--1313, 2015.

\bibitem{somers2016human}
T.~Somers and G.~A. Hollinger, ``Human--robot planning and learning for marine
  data collection,'' \emph{Autonomous Robots}, vol.~40, no.~7, pp. 1123--1137,
  2016.

\bibitem{chu2005gaussian}
W.~Chu, Z.~Ghahramani, and C.~K. Williams, ``Gaussian processes for ordinal
  regression.'' \emph{Journal of machine learning research}, vol.~6, no.~7,
  2005.

\bibitem{sui2018advancements}
Y.~Sui, M.~Zoghi, K.~Hofmann, and Y.~Yue, ``Advancements in dueling bandits,''
  in \emph{IJCAI}, 2018, pp. 5502--5510.

\bibitem{joachims2005accurately}
T.~Joachims, L.~A. Granka, B.~Pan, H.~Hembrooke, and G.~Gay, ``Accurately
  interpreting clickthrough data as implicit feedback,'' in \emph{SIGIR},
  vol.~5, 2005, pp. 154--161.

\bibitem{kandasamy2015high}
K.~Kandasamy, J.~Schneider, and B.~P{\'o}czos, ``High dimensional bayesian
  optimisation and bandits via additive models,'' in \emph{International
  conference on machine learning}.\hskip 1em plus 0.5em minus 0.4em\relax PMLR,
  2015, pp. 295--304.

\bibitem{wang2013bayesian}
Z.~Wang, M.~Zoghi, F.~Hutter, D.~Matheson, N.~De~Freitas \emph{et~al.},
  ``Bayesian optimization in high dimensions via random embeddings.'' in
  \emph{IJCAI}, 2013, pp. 1778--1784.

\bibitem{thompson1933likelihood}
W.~R. Thompson, ``On the likelihood that one unknown probability exceeds
  another in view of the evidence of two samples,'' \emph{Biometrika}, vol.~25,
  no. 3/4, pp. 285--294, 1933.

\bibitem{lai1985asymptotically}
T.~L. Lai and H.~Robbins, ``Asymptotically efficient adaptive allocation
  rules,'' \emph{Advances in applied mathematics}, vol.~6, no.~1, pp. 4--22,
  1985.

\bibitem{zoghi2014relative}
M.~Zoghi, S.~Whiteson, R.~Munos, and M.~Rijke, ``Relative upper confidence
  bound for the k-armed dueling bandit problem,'' in \emph{International
  conference on machine learning}.\hskip 1em plus 0.5em minus 0.4em\relax PMLR,
  2014, pp. 10--18.

\bibitem{kirschner2019adaptive}
J.~Kirschner, M.~Mutny, N.~Hiller, R.~Ischebeck, and A.~Krause, ``Adaptive and
  safe bayesian optimization in high dimensions via one-dimensional
  subspaces,'' in \emph{International Conference on Machine Learning}.\hskip
  1em plus 0.5em minus 0.4em\relax PMLR, 2019, pp. 3429--3438.

\bibitem{ailon2012active}
N.~Ailon, ``An active learning algorithm for ranking from pairwise preferences
  with an almost optimal query complexity.'' \emph{Journal of Machine Learning
  Research}, vol.~13, no.~1, 2012.

\bibitem{cosner2022safety}
R.~Cosner, M.~Tucker, A.~Taylor, K.~Li, T.~Molnar, W.~Ubelacker, A.~Alan,
  G.~Orosz, Y.~Yue, and A.~Ames, ``Safety-aware preference-based learning for
  safety-critical control,'' in \emph{Learning for Dynamics and Control
  Conference}.\hskip 1em plus 0.5em minus 0.4em\relax PMLR, 2022, pp.
  1020--1033.

\bibitem{li2022natural}
K.~Li, M.~Tucker, R.~Gehlhar, Y.~Yue, and A.~D. Ames, ``Natural multicontact
  walking for robotic assistive devices via musculoskeletal models and hybrid
  zero dynamics,'' \emph{IEEE Robotics and Automation Letters}, vol.~7, no.~2,
  pp. 4283--4290, 2022.

\end{thebibliography}

\newpage
\section{Appendix}
\subsection{Convex Program for Computing the MAP estimate}

The convex program to solve for $\bm{\util}_{\text{MAP}}$ is constructed as:
\begin{align*}
    \bm{\util}_{\text{MAP}} &= \argmin_{\utilvec \in \R^{|\actions|}} ~\S(\utilvec),
\end{align*}
with the first derivative terms of $\S(U)$ being:

\small{
\begin{align*}
    \frac{\partial -\ln \P(\act_{k1} \succ \act_{k2} \mid \utilvec(\act_{k1}), \utilvec(\act_{k2}))}{\partial \utilvec(\act_i)} &= \frac{-s_k(\act_i)}{c_p} \frac{\dot{g}(z_k)}{g(z_k)}, \\
    \frac{\partial -\ln \P(\bar{\act}_l \succ \act_{l} \mid \utilvec(\bar{\act}_l), \utilvec(\act_{l}))}{\partial \utilvec(\act_i)} &= \frac{-s_l(\act_i)}{c_c} \frac{\dot{g}(z_l)}{g(z_l)}, \\
    \frac{\partial -\ln \P( (\act_m,o_m) \mid \utilvec(\act_m))}{\partial \utilvec(\act_i)} &= \frac{1}{c_o} \frac{\dot{g}(z_{m1}) - \dot{g}(z_{m2})}{g(z_{m1}) - g(z_{m2})},
\end{align*}
}
\normalsize{
and the second derivative terms: 
}
\small{
\begin{align*}
    \frac{\partial ^2 -\ln \P(\act_{k1} \succ \act_{k2} \mid \utilvec(\act_{k1}), \utilvec(\act_{k2}))}{\partial \utilvec(\act_i)\utilvec(\act_j)} &= \dots\\
     \frac{s_k(\act_i)s_k(\act_j)}{c_p^2} & \left( \frac{\dot{g}(z_k)^2}{g(z_k)^2} - \frac{\ddot{g}(z_k)}{g(z_k)} \right), \\
    \frac{\partial^2 -\ln \P(\bar{\act}_l \succ \act_{l} \mid \utilvec(\bar{\act}_l), \utilvec(\act_{l}))}{\partial \utilvec(\act_i)\utilvec(\act_j)} &= \dots\\
    \frac{s_l(\act_i)s_l(\act_j)}{c_c^2} & \left( \frac{\dot{g}(z_l)^2}{g(z_l)^2} - \frac{\ddot{g}(z_l)}{g(z_l)} \right), \\
    \frac{\partial^2 -\ln \P( (\act_m,o_m) \mid \utilvec(\act_m))}{\partial \utilvec(\act_i)\utilvec(\act_j)} &= \dots \\
    \frac{1}{c_o^2} \Bigg( \bigg( \frac{\dot{g}(z_{m1}) - \dot{g}(z_{m2})}{g(z_{m1}) - g(z_{m2})}  \bigg)^2   -  \bigg( & \frac{\ddot{g}(z_{m1}) - \ddot{g}(z_{m2})}{g(z_{m1}) - g(z_{m2})}  \bigg) \Bigg),
\end{align*}
}
\normalsize{where $g: \R \to (0,1)$ represents any link function with first derivative $\dot{g}(\cdot)$ and second derivative $\ddot{g}(\cdot)$. The link function terms are defined as:
}
\small{
\begin{align*}
    &z_k = \left(\frac{\utilvec({\act_{k1}}) - \utilvec({\act_{k2}}) }{ c_p } \right), \quad 
    &&z_l = \left(\frac{\utilvec({\bar{\act}_{l}}) - \utilvec({\act_{l}}) }{ c_c } \right), \\
    &z_{m1} = \frac{b_{o_m} - \utilvec(\act_m)}{c_o}, \quad
    &&z_{m2} = \frac{b_{o_m-1} - \utilvec(\act_m)}{c_o}.
\end{align*}
}
\normalsize{
Lastly, the indicator functions are defined as:
\begin{align*}
    s_k(\act) = \begin{cases} +1 & \act = \act_{k1} \\ -1 & \act = \act_{k2} \\ 0 & \text{otherwise} \end{cases}, \quad 
    s_l(\act) = \begin{cases} +1 & \act = \bar{\act}_l \\ -1 & \act = \act_{l} \\ 0 & \text{otherwise} \end{cases}.
\end{align*}
}
\normalsize

\subsection{User Feedback Collection}
The process of querying the user for feedback is detailed in Alg. \ref{alg: feedback}.

\begin{algorithm}[t]
\caption{Obtain User Feedback (During Iteration $i$)}
\begin{small}
\hspace*{\algorithmicindent} 
\begin{algorithmic}[1]
\State $n := \text{num. of samples per iter.}$
\State $b := \text{num. of past samples in user's memory buffer}$
\State ${{n+b}\choose 2} :=$ number of pairwise comparisons per iteration
\For{$j = 1,\dots, {{n+b}\choose 2}$}
\If{User prefers first action ($\act_{j1}$) in comparison $j$}
    \State Record pairwise preference as $p_j = (\act_{j1} \succ \act_{j2})$
\ElsIf{User prefers second action ($\act_{j2}$) in comparison $j$}
    \State Record pairwise preference as $p_j = (\act_{j2} \succ \act_{j1})$
\ElsIf{No preference}
    \State $p_j = \emptyset$
\EndIf
    \State Append feedback to dataset: $\prefdata = \prefdata \cup p_j$
\EndFor
\For{$j = 1,\dots,n$}
\If{User has a suggestion regarding sampled action $\act_i^j$}
    \State Record coactive action as $\bar{\act}_i^j$
    \State Append feedback to dataset: $\coacdata = \coacdata \cup (\bar{\act}_i^j \succ \act_i^j)$
\EndIf
\EndFor
\For{$j = 1,\dots,n$}
\If{User has an ordinal label regarding sampled action $\act_i^j$}
    \State Record ordinal label as $o_i^j$
    \State Append feedback to dataset: $\coacdata = \coacdata \cup (\act_i^j,o_i^j)$
\EndIf
\EndFor
\end{algorithmic}
\end{small}
\label{alg: feedback}
\end{algorithm}

\subsection{PseudoCode for POLAR Toolbox}
Pseudocode outlining the general learning procedure enacted by the POLAR toolbox is outlined in Alg. \ref{alg: polar}.

\begin{algorithm}[t]
\caption{POLAR}
\begin{small}
\hspace*{\algorithmicindent} 
\begin{algorithmic}[1]
\State $\{\prefdata, \coacdata, \orddata, \vis\} = \emptyset$
\For{$i = 1, \dots, N$}
\If{$ i == 1$}
\State $\act_{i}:=$ set of $n$ random actions
\Else
\If{Regret Minimization}
    \If{Use Subset}
    \State Generate $L :=$ random line through $a^*_{i-1}$
    \State Construct $\sub := L \cup \vis$
    \State Approximate $P(\utilvec_{\sub} | \prefdata,\coacdata,\orddata)$ as $\N(\mu_{\sub},\Sigma_{\sub})$
    \State Obtain $\act_{i}$ using TS with  $\N(\mu_{\sub},\Sigma_{\sub})$
    \Else
    \State Obtain $\act_{i}$ using TS with  $\N(\mu_{\actions},\Sigma_{\actions})$
    \EndIf
\ElsIf{Active Learning} 
    \If{Use Subset}
    \State Obtain $\act_{i}$ using IG with  $\N(\mu_{\sub},\Sigma_{\sub})$
    \Else
    \State Obtain $\act_{i}$ using IG with  $\N(\mu_{\actions},\Sigma_{\actions})$
    \EndIf
\EndIf
\EndIf
\State Execute $\act_i$ and update visited actions $\vis = \vis \cup \act_{i}$
\State Store last $b$ actions of $\act_i$ as the buffer actions $\buffer$
\State Update $\prefdata,\coacdata,\orddata$ using Alg. \ref{alg: feedback}
\If{Regret Minimization}
\If{Use Subset}
\State Approximate $P(\utilvec_{\vis} \mid \prefdata,\coacdata,\orddata)$ as $\N(\mu_{\vis}, \Sigma_{\vis})$
\State Update $a_i^* := \argmin_{a \in \vis} \mu_{\vis}(a)$
\Else
\State Approximate $P(\utilvec_{\actions} \mid \prefdata,\coacdata,\orddata)$ as $\N(\mu_{\actions}, \Sigma_{\actions})$
\State Update $a_i^* := \argmin_{a \in \actions} \mu_{\actions}(a)$
\EndIf
\ElsIf{Active Learning}
\If{Use Subset}
    \State Update $\act_{\text{rand}}$ with $R$ random actions
    \State Construct $\sub := \act_{\text{rand}} \cup \vis$
    \State Approximate $P(\utilvec_{\sub} \mid \prefdata,\coacdata,\orddata)$ as $\N(\mu_{\sub},\Sigma_{\sub})$
    \Else
    \State Approximate $P(\utilvec_{\actions} \mid \prefdata,\coacdata,\orddata)$ as $\N(\mu_{\actions},\Sigma_{\actions})$
\EndIf
\EndIf
\EndFor
\end{algorithmic}
\end{small}
\label{alg: polar}
\end{algorithm}

\subsection{Details on Computing $\hat{\utilvec}_{\sub}$.}
Computing $\hat{\utilvec}_{\sub}$ has some slight modifications which we will present here. First, the likelihood functions are calculated only considering $\act \in \sub$. Second, the Gaussian prior is redefined over $\utilvec_{\sub}$:
\begin{align*}
    \P(\utilvec_{\sub}) = \frac{1}{(2\pi)^{\frac{|\sub|}{2}} | \Sigma_{\sub}^{\text{pr}}|^{\frac{1}{2}}} \exp \left( -\frac{1}{2}\utilvec_{\sub}^{\top} (\Sigma_{\sub}^{\text{pr}})^{-1} \utilvec_{\sub} \right),
\end{align*}
where $\Sigma_{\sub}^{\text{pr}} \in \R^{|\sub| \times |\sub|}$ is the prior covariance matrix with $[\Sigma_{\sub}^{\text{pr}}]_{ij} = \mathcal{K}(\act^i_{\sub},\act^j_{\sub})$ for the restricted set of actions $\act_{\sub}$ in $\sub$. Using these restricted likelihood terms and Gaussian prior, the posterior over $\sub$ is modeled as:
\begin{align*}
    &P(\utilvec_{\sub} | \data) \propto P(\prefdata | \utilvec_{\sub})P(\coacdata | \utilvec_{\sub})P(\orddata | \utilvec_{\sub})P(\utilvec_{\sub}),
\end{align*}
and again approximated using the same procedure presented in Sec. \ref{Sec: modeling} to obtain $\N(\mu_{\sub}, \Sigma_{\sub})$, where $\mu_{\sub}$ and $\Sigma_{\sub}$ denote the posterior mean and covariance over the subset $\sub$. Lastly, $\hat{\utilvec}_{\sub}$ is computed as the solution to the minimization: 
\begin{align*}
    \hat{\utilvec}_{\sub} = \argmin_{\utilvec \in \R^{|\sub| }} ~\S(\utilvec).
\end{align*}
\end{document}